\documentclass[sigconf, nonacm, review=false]{acmart}
\usepackage{multirow, dsfont, enumitem}

\AtBeginDocument{%
  \providecommand\BibTeX{{%
    \normalfont B\kern-0.5em{\scshape i\kern-0.25em b}\kern-0.8em\TeX}}}

\setcopyright{acmcopyright}
\copyrightyear{2022}
\acmYear{2022}
\acmDOI{XXXXXXX.XXXXXXX}

\acmConference[KDD '22]{28th ACM SIGKDD Conference on Knowledge Discovery and Data Mining}{August 14--18, 2022}{Washington DC, USA}

\begin{document}
\newcommand{\psold}{PinnerSage}
\newcommand{\psnew}{\textsc{PinnerFormer}}
\newcommand{\longpsold}{PinnerSage}
\newcommand{\longpsnew}{\textsc{PinnerFormer}}
\newcommand{\jure}[1]{{{\textcolor{red}{[Jure: #1]}}}}
\title{\longpsnew{}: Sequence Modeling for User Representation at Pinterest}

\author{Nikil Pancha}
\email{npancha@pinterest.com}
\affiliation{%
  \institution{Pinterest}
  \city{San Francisco}
  \country{USA}
}

\author{Andrew Zhai}
\email{andrew@pinterest.com}
\affiliation{%
  \institution{Pinterest}
  \city{San Francisco}
  \country{USA}
}

\author{Jure Leskovec}
\email{jure@cs.stanford.edu}
\affiliation{%
  \institution{Stanford University}
  \country{USA}
}

\author{Charles Rosenberg}
\email{crosenberg@pinterest.com}
\affiliation{%
  \institution{Pinterest}
  \city{San Francisco}
  \country{USA}
}

\begin{abstract}
  Sequential models have become increasingly popular in powering personalized recommendation systems over the past several years. These approaches traditionally model a user's actions on a website as a sequence to predict the user's next action. While theoretically simplistic, these models are quite challenging to deploy in production, commonly requiring streaming infrastructure to reflect the latest user activity and potentially managing mutable data for encoding a user's hidden state. Here we introduce \longpsnew{}, a user representation trained to predict a user's future long-term engagement using a sequential model of a user's recent actions. Unlike prior approaches, we adapt our modeling to a batch infrastructure via our new dense all-action loss, modeling long-term future actions instead of next action prediction. We show that by doing so, we significantly close the gap between batch user embeddings that are generated once a day and realtime user embeddings generated whenever a user takes an action. We describe our design decisions via extensive offline experimentation and ablations and validate the efficacy of our approach in A/B experiments showing substantial improvements in Pinterest's user retention and engagement when comparing \longpsnew{} against our previous user representation. \longpsnew{} is deployed in production as of Fall 2021. 
\end{abstract}

\ccsdesc[500]{Information systems~Recommender systems}
\ccsdesc[500]{Information systems~Personalization}

\keywords{Representation Learning, Multi-Task Learning, Personalization, Recommender Systems}

\maketitle
\section{Introduction}
Over 400M users use Pinterest each month to discover ideas and inspiration from our content corpus of billions of Pins. A Pin starts with an image and often includes text, a web link, and a board that connects the individual Pin to a user curated collection of pins. Inspiration is the key to Pinterest and facilitated mainly through our search and recommendation systems, allowing users to find content through (a) Homefeed, our personalized recommendation product, (b) Related Pins, recommendations contextual to a query Pin, and (c) Search, recommendations relevant to a user text query. Users give feedback through interactions such as saving Pins to boards (Repin), clicking through to the underlying link, zooming in on one Pin (close-up), hiding irrelevant content, and more. To achieve our mission of bringing everyone the inspiration to create a life they love, we need to personalize our content with our user's interests and context, taking into consideration feedback a user has given on their Pinterest journey; i.e., we need a strong representation of our users.

Learning user embeddings (representations) has become an increasingly popular method of improving recommendations.
Such embeddings have been adopted to power ranking and candidate generation in industry, and are used to power personalized recommendations across YouTube \cite{covington2016deep}, Google Play \cite{yang2020mixed}, Airbnb search \cite{grbovic2018realtime}, JD.com search \cite{zhang2020personalized}, Alibaba \cite{li2019multiinterest, pi2019practice}, and more.
In addition to work on learning personalized embeddings, there is a body of work focused on directly building ranking models using sequential information \cite{chen2019behavior, zhou2019deep, pi2019practice, pi2020searchbased}, enabling personalization of recommendations based on a user's recent engagement.

User behavior on websites tends to be sequential in nature; actions can be ordered by the time at which they are taken, which naturally leads to sequential modeling methods.
Various methods have been proposed to predict future engagement based on users' sequences of historical interactions \cite{chen2021topk, sun2019bert4rec, hidasi2018recurrent, rendle2010factorizing, pal2020pinnersage}.
More recent works have applied various deep learning models, including recurrent neural networks (RNNs) and transformers for such sequential recommendations and obtained promising results \cite{chen2021topk, sun2019bert4rec, hidasi2018recurrent, kang2018self, donkers2017sequential, wu2017recurrent, yuan2019convolutional}.
Sequential models traditionally focus on a realtime setting, aiming to predict a user's next action or engagement from all actions leading up to that point.

In practice, there are two key challenges in deploying existing sequence modeling approaches to large web-scale applications: (a) cost of computation, and (b) infrastructure complexity.
Existing sequence modeling approaches broadly fall into two categories: stateless models, and stateful models.
Stateless models may have high computational cost, as an embedding must be computed from scratch after every action a user takes, while stateful models require robust reliable streaming infrastructure to handle potential errors or data corruption in the model's state for a given user \cite{pi2019practice}.

Here we present \longpsnew{}, an end-to-end learned user representation that has been deployed in production at Pinterest.
Similar to prior work on sequential user modeling, \longpsnew{} directly learns a representation based on a user's past pin engagement.
We propose a dense all action loss, which allows our embedding to capture a user's longer-term interests, rather than only predicting the next action.
This allows our embedding to be computed in an offline, batch setting, and simplifies infrastructure considerably.

We also address the infrastructure complexity challenge that at Pinterest manifests in the following way: there are tens of ranking models that could benefit from personalization, but developing a custom solution for each one is not scalable.
Rather than producing one user embedding for each model (which would increase complexity), we choose to invest in developing a single high quality user embedding that can be used for many downstream tasks.
Although performance on a specific task may be sacrificed in some cases, the complexity tradeoff makes a shared embedding favorable for the majority of use cases.

We evaluate \longpsnew{} in offline as well as online A/B experiments. In offline experiments, we show that this training objective nearly halves the gap in performance between a model inferred daily and a model inferred in realtime, and reflects a user's longer-term interests better than other approaches.
Then, we demonstrate the utility of \psnew{} as a feature, demonstrating that it enables significant online gains when used as a feature in multiple ranking models across different domains. %

\section{Design Choices}
We begin by discussing key design choices in \longpsnew{}.

{\bf Design Choice 1: Single vs. multiple embeddings for a single user.}
Most approaches to generating user representations produce a single embedding \cite{chen2021topk, sun2019bert4rec, hidasi2018recurrent, kang2018self, donkers2017sequential, wu2017recurrent, yuan2019convolutional}, but some focus on learning a fixed or variable number of user embeddings \cite{li2019multiinterest, pal2020pinnersage, weston2013nonlinear, liu2019singlevectorenough}.
In our previous user representation, \longpsold{}, we made the decision to allow for a variable, potentially large number of embeddings, allowing the model to explicitly represent a user's varied interests \cite{pal2020pinnersage}.

Although using multiple embeddings allows for a model to more explicitly capture user interests and works well for retrieval, this can lead to issues when used in downstream models: storing 20+ 256d float16 embeddings in training data does not scale well, especially when datasets may contain billions of rows, as they do for ranking models.
Separately, this also increases cost of model training and inference; processing 5000+ floating point numbers can introduce nontrivial latency, especially if they are transformed before aggregation.
At training time, large examples can also increase the time taken to load data.
To avoid these issues, when using \longpsold{} in ranking models we would typically use a weighted aggregation of a user's embeddings as the final user representation.
Because we wish for \longpsnew{} to be easily usable as a feature, we produce a single embedding that captures a user's interests, allowing for painless use in downstream models.
In offline evaluations, we show that our single embedding is able to reflect a user's longer-term interests better than \longpsold{}, while requiring a fraction of the storage.

\textbf{Design Choice 2: Real-time vs. Offline inference.} Most prior work on sequential user modeling focuses on models that operate in realtime or near realtime.
In practice, this leads to at least one of the following:
\begin{itemize}[topsep=2pt]
    \item \textbf{High computational cost:} for every action the user takes, the system must fetch all events in a user's history and frequently infer a potentially complex model
    \item \textbf{High infrastructure complexity:} a user's hidden state or embedding can be incrementally updated, but this requires a robust system to recover and warm up the model's state in the case of any data corruption \cite{pi2019practice}
\end{itemize}

On Pinterest, a user may take tens or hundreds of actions in a day, so a model that updates a user's embedding at most once each day only requires a small fraction of the computational resources of a comparably-sized realtime model.
In offline evaluations, we demonstrate that our loss formulation substantially decreases the gap between realtime and daily-inferred models, and in A/B experiments, we show \psnew{} greatly improves performance of downstream ranking models.
\begin{figure*}
    \centering%
    \includegraphics[width=0.9\textwidth]{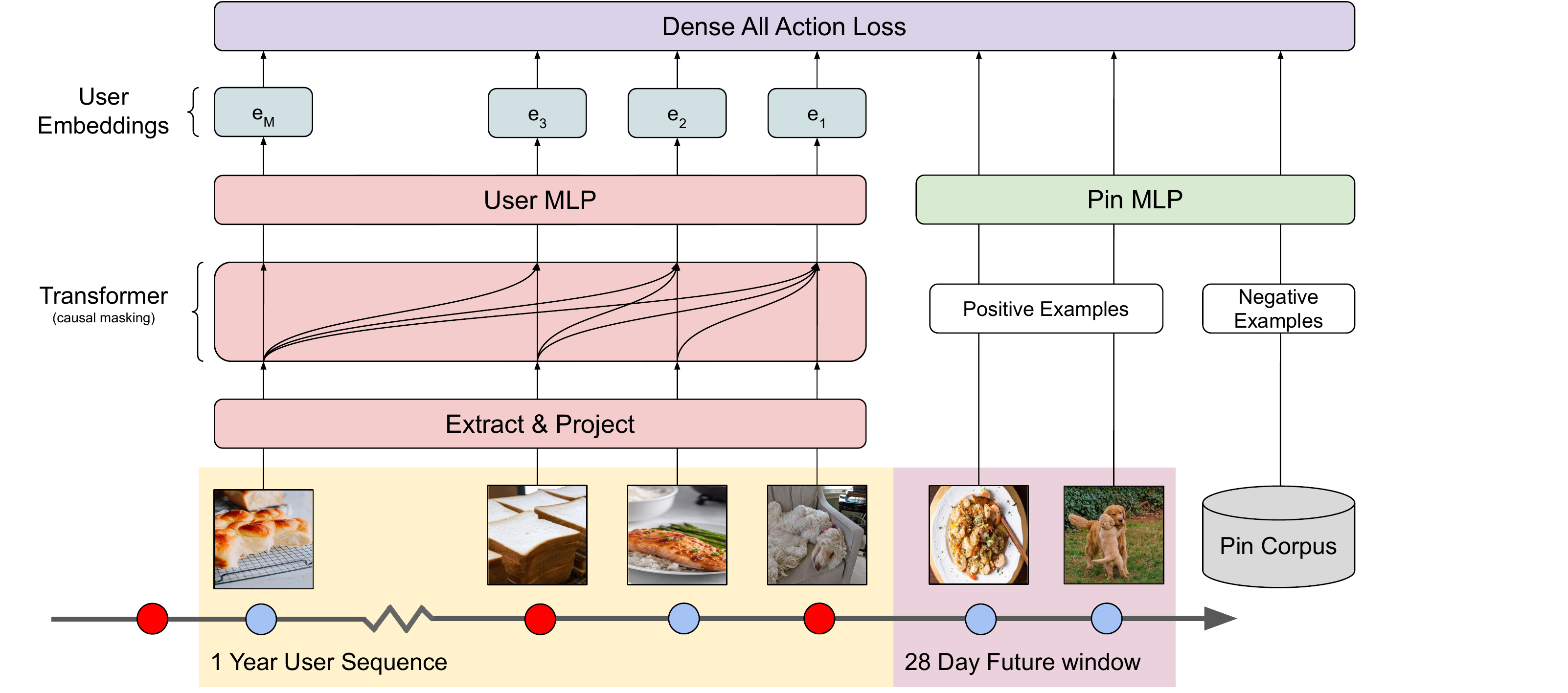}
    \caption{Overview of \psnew{} architecture. Features are passed through a transformer with causal masking, and embeddings are returned at every time step. Note that the training window (28d above) exceeds our future evaluation objective window (14d)}
    \label{fig:e2e_model}
\end{figure*}
\section{Our Approach: \longpsnew{}}
In this section, we present \longpsnew{}, which has been used in production at Pinterest since Fall 2021, describing both our model (depicted in Figure \ref{fig:e2e_model}) and how it is deployed.

We begin with a corpus of Pins $\mathcal{P} = \{P_1, P_2, \ldots, P_N\}$, where $N$ is large, on the order of billions, and a set of users $\mathcal{U} = \{U_1, U_2, \ldots\}$, where $\lvert\mathcal{U}\rvert > 500\text{M}$.
For each Pin in the corpus, we have a PinSage \cite{ying2018graph} embedding $p_i \in \mathbb{R}^{256}$, which is an aggregation of visual, text, and engagement information for a Pin $P_i$.
For each user, we have a sequence of actions a user has taken on the site, $\mathcal{A}_U = \{A_1, A_2, \ldots, A_S\}$, ordered ascending by timestamp.
In this work, we limit this sequence of actions to users' engagements with Pins, including Pin saves, clicks, reactions, and comments over the past year.
Based on this assumption, an action can then be represented by a PinSage embedding, as well as some metadata about the action. %
In practice, $S$ may be very large for a given user, on the order of thousands or tens of thousands for some users, so we compute a user's embedding using their $M$ most recent actions.

Given these definitions, we aim to learn a user representation $f\colon \mathcal{U} \mapsto \mathbb{R}^d$, that is compatible with some Pin representation $g\colon \mathcal{P}\mapsto \mathbb{R}^d$ under cosine similarity.
We learn $f$ and $g$ jointly, using the sequence of actions $\mathcal{A}_U$ as the only input features to the model, and restricting to only the latest $M$.

In a user's complete action sequence, there may be many types of actions, some of which are good (e.g. a long click), and and some of which are neutral or negative (e.g. a hide or short click).
In this work, we focus learning representations to predict \textit{positive} engagement, which we define as a Pin save ("Repin"), a Pin close-up lasting over 10s ("Closeup"), or a long clickthrough (>10s) to the link underlying a Pin ("Click").
We only treat engagement on Homefeed as positive; on surfaces such as Search or Related Pins, the query provides substantial context, while on Homefeed, the user provides the primary context.

Our primary objective is to learn a model that is able to predict a user's \textit{positive} future engagement over a 14 day time window after the generation of their embedding, rather than a traditional sequence modeling task, where the embedding would only predict the next action taken.
In other words, our goal is to learn embeddings $u_i$ and $p_i$ such that if $d(u_k, p_i) < d(u_k, p_j)$, then $p_i$ is more likely to be positively engaged with by the user represented by $u_k$ than $p_j$ over the 14 days after $u_k$ is generated.
We choose this range of 14 days for tractability, and assume that actions a user takes over the course of two weeks sufficiently are representative of a user's longer-term interests.
Figure \ref{fig:e2e_model} illustrates the \longpsnew{} architecture, and below we expand on each component in more detail.

\subsection{Feature Encoding}
For each action in a user's sequence, we have a PinSage embedding (256-dimensional) \cite{ying2018graph} and metadata features: action type, surface, timestamp, and action duration.
We use small, learnable embedding tables to encode action type and surface, our two categorical features, and drop sequence elements with out of vocabulary terms for either of these two features.
We encode action duration with a single scalar value, $\log(duration)$.

To represent the time an action occurred, we use 2 derived values in addition to the raw absolute timestamp: the time since the latest action a user has taken, and the time gap between actions.
For each of these time features, we follow the common practice of encoding time using sine and cosine transformations with various periods in a manner similar to Time2vec \cite{kazemi2019time2vec}, but with $P$ fixed periods, rather than learned periods, and a logarithmic transformation of time, rather than a linear one.
This produces $2P+1$ features ($2P$ from periodic transofmrations of the timestamp).\footnote{Details are described in Section \ref{appendix:timestamp} of the reproducibility material}

All features are concatenated into a single vector, resulting in an input vector of dimension $D_{\text{in}}$.
The representation corresponding to action $A_i$ is denoted as $a_{i} \in \mathbb{R}^{D_{\text{in}}}$.

\subsection{Model Architecture}
In \longpsnew{}, we model the sequence of user actions using a transformer model architecture \cite{vaswani2017attention}.
We choose to use PreNorm residual connections, applying Layer Normalization before each block, as this approach has been shown to improve stability of training \cite{nguyen2019transformers, wang2019learning}.
We first construct the input matrix $A = \begin{pmatrix} a_{T} &\cdots & a_{T - M + 1} \end{pmatrix}^\intercal \in \mathbb{R}^{M \times D_{\text{in}}}$ using the $M$ actions leading up to action $A_{T+1}$ as the user's sequence.
Then, we project these to the transformer's hidden dimension, add a fully learnable positional encoding, and apply a standard transformer consisting of alternating feedforward network (FFN) and multi-head self attention (MHSA) blocks.
The output of the transformer at every position is passed through a small MLP and $L_2$ normalized, resulting in a set of embeddings $E = \begin{pmatrix} e_{1} & \cdots & e_{M} \end{pmatrix}^\intercal \in \mathbb{R}^{S \times D}$, where $D$ is the final embedding dimension.\footnote{More precise equations are provided in Section \ref{appendix:model-arch} of the reproducibility material}

To represent Pins, we learn an MLP that takes only PinSage as an input, and $L_2$ normalize the output embeddings.
We found that using $L_2$ normalized embeddings to represent both users and Pins leads to the most stable training, and does not sacrifice offline performance.

\subsection{Metric Learning}
To train our representation, we need pairs $\{(u_1, p_1), \ldots, (u_B, p_B)\}$ consisting of user embeddings and target Pin embeddings, where both users and Pins may be repeated.
We choose to not use explicit negative examples in this work (i.e. we do not have loss terms for negative engagement, such as hides).
In designing our model, there are several considerations: 
\begin{enumerate}[label=(\alph*)]
    \item \label{item:metric:pairs} How do we choose these pairs?
    \item \label{item:metric:negatives} For a given $(u_i, p_i)$ pair, how do we select negative examples?
    \item \label{item:metric:loss} Given a $(u_i, p_i)$ pair and a set of negative examples, how do we compute the loss?
\end{enumerate}
We first describe \ref{item:metric:negatives} and \ref{item:metric:loss}, then in Section \ref{sec:training-objective} elaborate on \ref{item:metric:pairs}.
\subsubsection{Negative Selection}
We consider two sources of negative examples: in-batch negatives and random negatives.
When selecting in-batch negatives for a given user, we choose all positive examples within the batch as negatives, masking pins that have positive engagement for that user.
This approach is efficient and simple, but can lead to demotion of popular Pins if implemented naively, as engaging Pins are more likely to appear as negatives than less engaging ones.
Another downside to in-batch negatives is that the distribution of negative examples is different from the true underlying distribution of Pins used for retrieval, leading to a discrepancy between training and serving.
The second source of negatives are those uniformly sampled from the corpus of all Pins we might surface to Homefeed, but using these alone can lead to model collapse, as the negatives may be too easy.
A third option we consider is combining both random and in-batch negatives to take advantage of the unique characteristics of both by merging the in-batch and random negative pools into a single one, which contains a combination of in-batch and random negatives \cite{yang2020mixed}.

In practice, a larger negative pool can increase the quality of the learned embeddings, so we gather negative examples from across all GPUs used in training, choosing the largest possible pool that can comfortably fit in GPU memory.

\subsubsection{Loss Function}
After choosing the source of negative examples, we can produce a set of negative embeddings $\{n_1, \ldots, n_N\}$ for a given pair of user and positive embeddings $(u_i, p_i)$.
We compute a loss for each pair, and then compute a weighted average such that each user in the batch on a given GPU is given equal weight.

The loss function we have found to work best is sampled softmax with a logQ correction \cite{yi2019sampling_bias_corrected, bengio2008adaptive}, where we apply a correction to each logit based on the probability that a given negative appears in the batch. We also learn a temperature $\tau \in [0.01, \infty)$, constraining the lower bound for stability.
If we let $s(u, p) = \langle u, p \rangle / \tau$, a sampled softmax loss without sample probability correction would be defined as follows:
\begin{equation}
    \mathcal{L}(u_i, p_i) = -\log \left( \frac{e^{s(u_i, p_i)}}{e^{s(u_i, p_i)} + \sum_{j=1}^{N} e^{s(u_i, n_j)}}\right)
\end{equation}
When negatives are not uniformly distributed, A correction term $Q_{i}(v) = P(\text{Pin}\ v\ \text{in batch} \mid \text{User}\ U_i\ \text{in batch})$ should be applied to correct for sampling bias, where $v$ may be a positive or negative example.
The softmax loss with sample probability correction for a single pair is then defined as follows:
\begin{equation}
    \mathcal{L}(u_i, p_i) = -\log \left( \frac{e^{s(u_i, u_i) -\log (Q_{i}(p_i))}}{e^{s(u_i, u_i) - \log(Q_{i}(p_i))} + \sum_{j=1}^{N} e^{s(u_i, n_j) - \log(Q_{i}(n_j))}}\right)
\end{equation}
For simplicity, we approximate $Q$ using a count-min sketch \cite{cormode2005improved}.

\subsection{Training Objective}\label{sec:training-objective}
\begin{figure}
    \centering
    \includegraphics[width=\columnwidth]{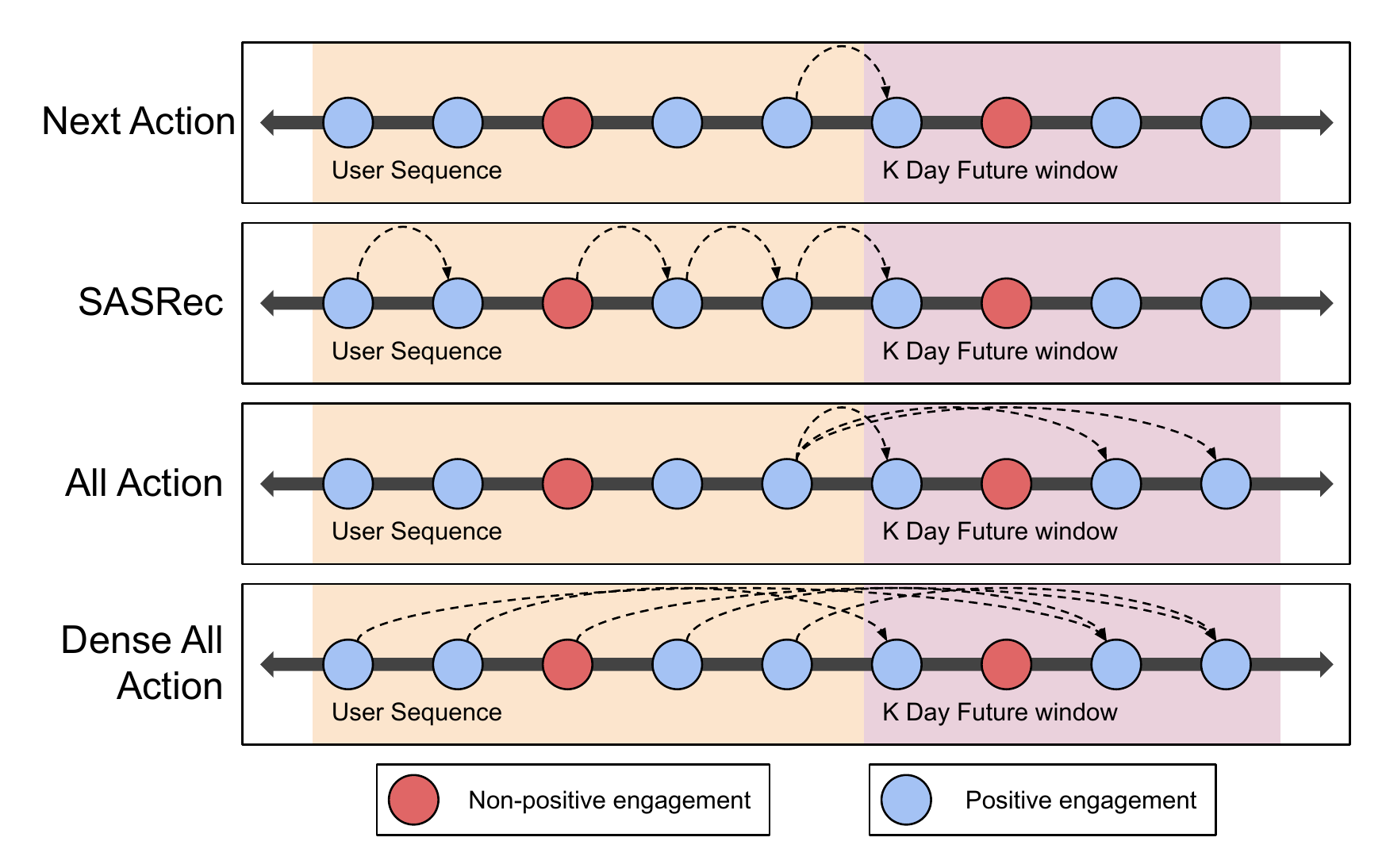}
    \caption{Four explored training objectives. Blue circles represent embeddings corresponding to actions considered positive, while red circles represent embeddings corresponding to actions considered non-positive (but not necessarily explicitly negative). The exact pairings in the dense all action loss are sampled, so this is simply one potential materialization. Note we do not attempt to predict non-positive examples}
    \label{fig:training-objectives}
\end{figure}

Given our loss function, we address the question of how to select pairs $(u_i, p_i)$.
There are three forms of positive engagement our model should able to predict: Repins, Closeups, and Clicks.
Each of these actions has value, but rather than learning task-specific heads as is common in multi-task learning literature \cite{liu2015representation, bell2020groknet}, we choose to train a single embedding in a multi-task manner, directly learning an embedding that can effectively retrieve different types of positive engagement.
We do not explicitly weight different engagement differently in our loss computation function.
The four training objectives we consider are described below, and are depicted in Figure \ref{fig:training-objectives}.

\subsubsection{Next Action Prediction} \label{sssec:nextactionpred}
The naive objective for a sequence modeling task is next action prediction, in which we predict $A_{T+1}$ given the user sequence $\{A_{T}, A_{T-1}, \ldots, A_{T - M + 1}\}$ (if $A_{T+1}$ is a positive engagement).
This objective is intuitive for a realtime sequence model because, in the online setting, $A_T$ will always be the most recent action a user has taken.
SASRec \cite{kang2018self} extends this simple training objective by aiming to predict the next action at every step in the model, rather than only predicting the most recent positive action.
We slightly modify this in our experiments, only allowing positive actions to contribute to the model's loss.

Unlike these traditional objectives, we do not aim to predict a user's next immediate action; instead, we infer our user embeddings daily and aim to capture longer-term interests of a user.
To do so we introduce two alternate training objectives.

\subsubsection{All Action Prediction} Based on the observation that we don't solely wish to predict the next action a user will take, we construct a na\"ive training objective that predicts \textit{all} actions a user will take over the next $K$ days using $e_1$, the final user embedding.\footnote{\label{foot:train-eval-window}$K$ may not equal 14: we fix our evaluation objective to a 14 day window, but training on the same window may not optimize performance, as shown in \ref{sssec:training-objective}}
Assuming a user has positive engagement in actions $T + 3$, $T + 8$, and $T + 12$, all of which fall within a $K$ day window of $T$, we aim to predict all 3 actions: $A_{T+3}$, $A_{T+8}$, $A_{T+12}$ from the user sequence $\{a_{T}, a_{T-1}, \ldots, a_{T - S + 1}\}$.
This objective forces the model to learn longer-term interests, rather than focusing solely on the next action a user will take, which should decrease the effect of staleness that comes from daily offline inference.
For computational tractability, we randomly sample up to 32 actions per user in this $K$ day time window.
\begin{figure}[t]
    \centering %
    \includegraphics[width=\columnwidth]{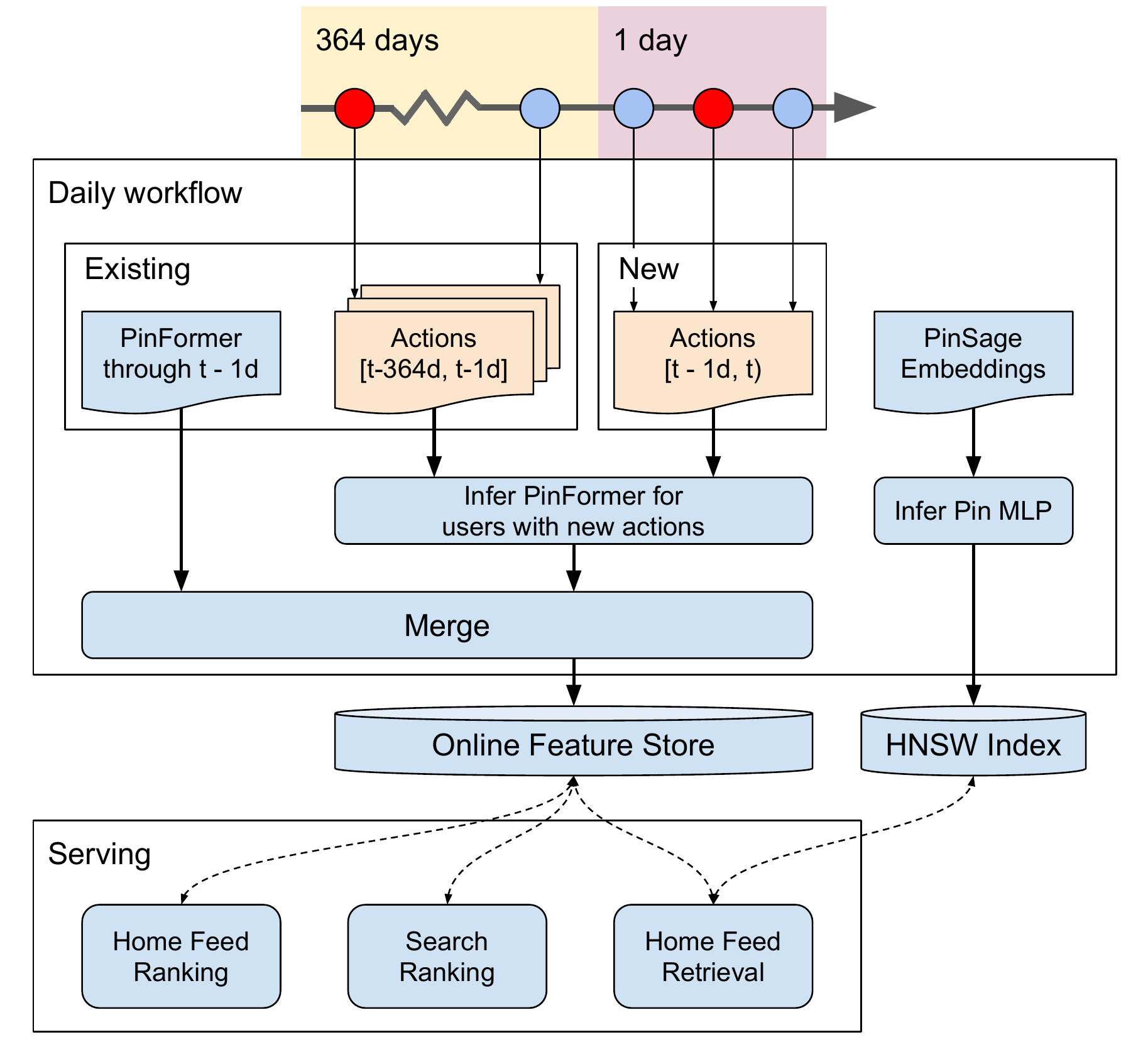}
    \caption{\psnew{} is inferred incrementally. We only compute embeddings for users who have engaged on Pinterest in the past day, then merge the new embeddings with the previous set of embeddings, falling back to the old ones if the new ones are missing}
    \label{fig:incr-infer}
\end{figure}

\subsubsection{Dense All Action Prediction}
To further improve the signal provided by each batch, we draw inspiration from SASRec \cite{kang2018self} to modify the all action prediction objective.
Rather than predicting actions over a $K$ day window using only $e_1$, the most recent user embedding, we instead select a set of random indices, $\{s_i\}$, and for each $e_{s_i}$, aim to predict a randomly selected positive action from the set of all positive actions over the next $K$ days.
To ensure this approach learns from the ordering of the data, we apply causal masking to the transformer's self-attention block, so each action may only attend to past or present actions, but not future actions.
We observe this masking substantially improves model performance on this task.
To decrease memory usage, we do not aim to predict all positive actions, and instead only aim to predict one positive action for each $e_{s_i}$.\footnote{Observe that without any subsampling, assuming a batch size of 128, and 32 sampled positives, and maximum sequence length $M=256$, this could generate up to $128 \cdot 256 \cdot 32=1048576$ pairs for softmax computation per GPU.}

\subsection{Dataset Design}
We make use of a compressed format to store training sequences.
One observation we make is that given a single user's timeline, many separate user sequences and positive examples can be constructed.
Given an entire user sequence $\mathcal{A}_U = \{A_1, \ldots, A_S\}$, and some maximum sequence length $M$, we can construct up to $S - M - 1$ training examples of length exactly equal to $M$ (assuming all actions are positive).
For example, the sequence $\{A_5, \ldots, A_{5 + M - 1}\}$ with positive engagements $\{A_{5 + M}, A_{7 + M}\}$ can be extracted from the complete timeline $\mathcal{A}_U$. %
One potential way to store this data would be to materialize all relevant sequences of length $M$ (or less) ahead of time, along with a corresponding set of future positive engagements for each sequence.
This runs into issues when experimenting with different sampling strategies, as tuning parameters would require regeneration of training data \textemdash a slow process.
To increase productivity, we instead store each user's sequence as a single row in our dataset, and sample examples on the fly during training.
This has a clear benefit of allowing for customized sampling during training, at the cost of decreasing shuffling of training data.

Specifically, there are several parameters we have tuned using this strategy, all of which can significantly impact the model's overall performance:
\begin{itemize}
    \item Maximum sequence length
    \item The fraction of possible user sequences that are sampled from a user's timeline
    \item Maximum number of sequences to sample per user
    \item Maximum number of positive examples to sample as labels for each sequence
\end{itemize}

\subsection{Model Serving}
As we focus on inference in an offline, batch setting for \psnew{}, we infer the model in a daily, incremental workflow, shown in Figure \ref{fig:incr-infer}.
This workflow generates new embeddings for users who have engaged with any Pin in the past day, merges them with the previous day's embeddings, then uploads them to a key-value feature store for online serving.
Because we only generate new embeddings for users who have engaged in the last day and run inference offline (with no latency constraints), we are able to use larger models than would otherwise be possible, which increases the information our embedding can capture.
In the case of any corruption in input features (for example, due to logging errors), we can easily run inference for all users whose embeddings have been updated since corruption and the next day's data will be correct, assuming the upstream data has been fixed.

Pin embeddings are inexpensive to compute, only requiring a small MLP transformation of an existing feature, so we generate them from scratch daily, then compile a HNSW \cite{malkov2018efficient} graph that can be queried online using the user embeddings saved in a feature store.

\section{Experiments and Results}
Here, first we compare \longpsnew{} with baselines, conduct ablations, and explore the gap in performance between realtime and daily inference through offline experiments. Then, we show considerable improvements over \longpsold{} in A/B experiments.
\subsection{Offline Evaluation Metrics}
The primary metric we use for evaluation is Recall@10. We select a 2 week period after training ends for evaluation, and evaluate on a disjoint set of users from those used for training.
Assuming the training dataset ends at time $t$, we compute embeddings at time $t$ for all users in the evaluation set, and then measure how well the embedding at time $t$ retrieves \textit{all} Pins a user will engage with from time $t$ to $t+14d$ from an index of 1M random Pins.
Assuming we have a set of users, $\mathcal{U}$, a set of positively engaged Pins $\mathcal{P}_U$ for each user $U$, and a random corpus of 1M Pins $\mathcal{N}$, we compute Recall@k (R@k) as follows:
\begin{align*}
    \mathrm{Recall@k}(U) &= \frac{1}{\lvert\mathcal{P}_U\rvert} \sum_{P \in \mathcal{P}_U} \mathds{1}\left\{\left\lvert\left\{ N \in \mathcal{N} \mid \mathrm{d}(U, P) \ge \mathrm{d}(U, N) \right\}\right\rvert < k\right\} \\
    \mathrm{Recall@k} &= \frac{1}{\lvert \mathcal{U} \rvert}\sum_{U\in \mathcal{U}} \mathrm{Recall@k}(U)
\end{align*}
Here, distance between a user and pin is defined by the Euclidean distance between the user's embedding and the pin's embedding.

We also observe two measures of diversity: (a) the entropy of the distribution of Interests (about 350 unique topics of Pins) associated with the top 50 retrieved results from an index of 1M Pins ("Interest Entropy@50"), and (b) what fraction of the index of 1M Pins accounts for 90\% of the top 10 retrieved results over a set of users ("P90 Coverage@10").
The former measures the diversity of the results retrieved for a specific user, whereas the latter represents the global diversity of retrieved results across all users.
Both are useful to observe; a simple baseline which memorizes popularity independent of user could have good performance by metric (a), but (b) will show a value very close to 0.0.

\subsection{Offline Results}
In this section we first compare \longpsnew{} with baselines, then investigate what aspects of the model lead to good performance.
\subsubsection{Comparison with Baselines}
\begin{table}
    \centering
  \caption{\longpsold{} (PS) vs \longpsnew{}. \psnew{} outperforms \longpsold{} on our engagement evaluation, even when \longpsold{} is evaluated by setting the user embedding to the closest cluster to the true positive embedding. 
  Higher interest entropy indicates more diverse results are retrieved per user, and higher coverage indicates that more unique results are retrieved over all users.
  }
  \label{tab:model-comparsion}
  \begin{tabular}{lrrr}
    \toprule
    Model & R@10 & \begin{tabular}{@{}c@{}}Interest\\Entropy@50\end{tabular} & P90 Coverage@10 \\
    \cmidrule(r){1-1}\cmidrule(l){2-4}
PS (5 clusters) & 0.026 & 1.69 & 0.130 \\
PS (20 clusters) & 0.046 & 2.10 & 0.133 \\
\psnew{} & 0.229 & 1.97 & 0.042 \\
    \bottomrule
  \end{tabular}
\end{table}
In our offline evaluation, we compare to the baseline of \longpsold{} \cite{pal2020pinnersage}, our previous, multi-embedding user representation, measuring recall based on an oracle evaluation to obtain an upper bound.
Specifically, given a fixed cutoff $c$, and a given positive, we choose the user's representation as the closest embedding to the positive among the top $c$ \longpsold{} embeddings.
We believe this establishes an approximate upper bound on the ability of those $c$ embeddings to predict engagement.
To compute diversity metrics, we do not adopt the oracle approach, and instead order results with round-robin blending: given some set of user embeddings (ordered by weight), each with some retrieved results, we take the first result from the first user embedding, the second from the second, and so on, returning to the first embedding after each has retrieved one result.

In Table \ref{tab:model-comparsion}, we show comparisons between \psnew{} and \longpsold{}, evaluated as described above.
Even when evaluating \psold{} with an oracle over the top 5 or 20 clusters, we see that the single \psnew{} embedding outperforms \longpsold{} in terms of retrieving content users are likely to engage with over a 14 day period.
Increasing the number of clusters used to retrieve results leads to more diversity in the results retrieved for a given user, which is an area \longpsold{} outperforms \psnew{} when using a sufficiently large number of clusters.
We also see that \longpsold{} retrieves more unique candidates from the index, but certain variants of \psnew{} achieve comparable levels of unique candidates while keeping engagement evaluation metrics higher, as seen in Table \ref{tab:softmax-spc}.

\subsubsection{Daily vs Realtime Inference}
\begin{table}
  \caption{Real-time vs. offline batch inference. Moving from realtime to batch inference drops Recall@10 by 13.9\% when training on SASRec objective, but only by 8.3\% when using a dense all action objective (\psnew{})}
  \label{tab:batch-realtime}
  \begin{tabular}{llrr}
    \toprule
    \begin{tabular}{@{}c@{}}Inference\\Frequency\end{tabular} & Model & R@10 & P90 Coverage@10 \\
    \cmidrule(r){1-2}\cmidrule(l){3-4}
\multirow{2}{*}{Once}  & SASRec & 0.198 & 0.048\\
& \psnew{} & 0.229 & 0.042 \\
\multirow{2}{*}{Daily}  & SASRec & 0.216 & 0.052\\
& \psnew{} & 0.243 & 0.043 \\
\multirow{2}{*}{Realtime}  & SASRec & 0.251 & 0.057 \\
& \psnew{} & 0.264 & 0.045 \\ %
    \bottomrule
  \end{tabular}
\end{table}
To quantify the drop in performance as inference frequency decreases, we compare two models that only vary in their training objective:
\begin{itemize}
    \item A model trained using the SASRec training objective, which directly predicts the next action a user will take\cite{kang2018self}. We replace binary cross-entropy with our sampled softmax loss, and separately weight the loss on $e_1$ equal to the loss on other positions, as we find this improves performance.\footnote{See Section \ref{appendix:sasrec} of reproducibility material for further justification of this change}
    \item \longpsnew{}, trained using the dense all action prediction objective with a 28d window
\end{itemize}

We then evaluate these at three different frequencies over the evaluation window $(t, t+14d]$:
\begin{itemize}
    \item \textbf{Once:} We use the single embedding predicted for a user at time $t$ to predict all of a user's positive actions over the 14 day window $(t, t+14d]$.
    \item \textbf{Daily:} We update a user's embeddings every day, predicting their actions in the interval $(x, x+1d]$ based on the embedding computed at $x-1d$, where $x \in \{t, t+1d, \ldots, t+13d\}$. This one day gap accounts for the delay between when an action is available in offline logs, and when it is uploaded to the feature store
    \item \textbf{Realtime:} We update embeddings after every action; i.e. we use the sequence of a user's $M$ actions preceding positive action to predict that positive action (for all positive actions in $(t, t+14d]$).
\end{itemize}
Note the daily and realtime settings are different from our primary evaluation.
Here, given a user's embedding at a point in time, we measure our ability to predict the \textit{specific} action a user will take, while our primary evaluation measures the ability of the embedding to capture a user's longer-term interests.

A realtime model is not practical to serve in production, as it would substantially increase inference cost over a batch model: some users may take tens or hundreds of actions per day, which translates to many times the cost of an offline model, even if using a shorter sequence.
We expect this realtime baseline to perform better than an offline, daily-computed model, but it helps quantify the opportunity cost of avoiding the realtime setting.

In Table \ref{tab:batch-realtime}, we also notice that the performance of \longpsnew{} increases as the inference frequency increases, once at start of eval, to once daily, to realtime.
Surprisingly, even in realtime \psnew{} outperforms a model trained to predict only the next engaged item.

This experiment also provides evidence that the dense all action prediction objective has the desired effect of decreasing the model's sensitivity to short-term variations, and instead learns more stable interests of a user: when moving from realtime to daily inference, and daily inference to inference only once, there is a smaller loss in performance when the model is trained on the dense all action objective (-8.3\%) than a next action prediction task (-13.9\%).

There is still a nontrivial gap between realtime performance and daily inference performance, but given improvements over our baseline of \longpsold{}, and the high cost and infra complexity of inferring \psnew{} in realtime, we view this as an acceptable tradeoff.

\subsubsection{Training Objective Selection}\label{sssec:training-objective}
\begin{table}
\centering
  \caption{Comparison of various training objectives. The dense all action objective maximizes Recall@10, and a 28 day future window performs significantly better than a 14d window.}
  \label{tab:tf-losses}
  \begin{tabular}{lrr}
    \toprule
    Training Objective & Recall@10 & P90 Coverage@10\\
    \cmidrule(r){1-1}\cmidrule(l){2-3}
Next Action             &  0.186   &   0.050 \\
SASRec (Softmax)         &  0.198   &   0.048 \\
All Action (28d)        &  0.224   &   0.028 \\
Dense All Action (14d)  &  0.223   &   0.043 \\
Dense All Action (28d)  &  0.229   &   0.042 \\
    \bottomrule
  \end{tabular}
\end{table}
In Table \ref{tab:tf-losses}, we observe that only training to predict the next action in training leads to lower Recall@10, but higher retrieved index coverage.
The lower Recall@10 can be explained because all action prediction tasks align better with the evaluation objective than next action prediction.
We believe we observe the higher index coverage for next action prediction because prediction of actions over a longer time frame is a harder task than only predicting the next action taken, so the Pin embedding learned may bias more towards retrieving more generally engaging content than retrieving content directly relevant to recent actions.
We have also observed that training on a 28 day future window for all action prediction yields better results than a 14 day window, even when the evaluation is fixed to a 14 day window.
We believe this can be explained as a consequence of having more labels for each user sequence, which can increase training efficiency.

The dense all action loss outperforms all action prediction on Recall@10 and global diversity.
The key difference between these two losses is that in the all action loss, gradients from all positive examples for a user will be backpropagated through the same user embedding, resulting in a larger averaging effect, as compared to the dense all action loss, where gradients all pass through different transformer outputs, and are only averaged together after passing into the transformer.

We also tried summing losses computed based on different training objectives together, but such configurations did not outperform any single-objective model.

\subsubsection{Sampled Softmax}
\begin{table}
    \centering
  \caption{Effect of negative pool and sample probability correction (SPC). SPC significantly improves Recall@10, at the cost of decreased global result diversity.}
  \label{tab:softmax-spc}
  \begin{tabular}{cccr}
    \toprule
    SPC & Negative Source & P90 Coverage@10 & Recall@10\\
    \cmidrule(r){1-2}\cmidrule(l){3-4}
N & random &                 0.002 &              0.136 \\
N & in-batch &                 0.163 &              0.071 \\
N & mixed &                 0.083 &              0.138 \\
Y  & random &                 0.001 &              0.139 \\
Y  & in-batch &                 0.119 &              0.167 \\
Y  & mixed &                 0.042 &              0.229 \\
    \bottomrule
  \end{tabular}
\end{table}

In Table \ref{tab:softmax-spc}, we compare performance of different settings for our softmax loss.
In all cases, we see that the presence of in-batch negatives increases diversity of retrieved results, but results in lower Recall@10 than mixed negatives.
When we train a model using random negatives, the model seems to collapse to retrieving very similar results for all users; when retrieving 10 out of 1M Pins for 100000 users, only 1000 Pins can account for 90\% of the retrieved results. This seems to indicate the model is failing to learn fine details regarding a user's interests, as it is retrieving very similar content for most users.

Overall, we see that sample probability correction does not increase Recall@10 on random negatives, which is expected: the probability that all negatives are in the batch should be equal in this case, as sampling is unbiased.
When including in-batch negatives (either alone, or in combination with random negatives), enabling sample probability correction increases recall while decreasing global diversity.
Given the large difference between in-batch and mixed negatives in terms of Recall@10, we choose to use mixed negatives with sample probability correction as our loss function, even though mixed negatives introduce slightly more complexity.

\subsubsection{Multi-task Learning}
\begin{table}
    \centering
    \caption{Performance on different action types vs training objective action type. The best performing objective for each task is bold, second best is underlined. Although multi-task does not perform as well as single-task learning, it performs better than any individual model on overall Recalll@10, and second best on all other tasks.}
    \label{tab:mtl-eval}
    \begin{tabular}{lcccc}
    \toprule
    & \multicolumn{4}{c}{Evaluation Task} \\
    \cmidrule(l){2-5}
    Training Objective  & 10s Closeup    & 10s Click & Repin & All \\
    \midrule
    10s Closeup        & \textbf{0.27} & 0.02 & 0.09 & \underline{0.17} \\
    10s Click           & 0.01 & \textbf{0.49} & 0.01 & 0.12 \\
    Repin               & 0.15 & 0.03 & \textbf{0.17} & 0.13 \\
    Multi-task          & \underline{0.23} & \underline{0.28} & \underline{0.13} & \textbf{0.23} \\
    \bottomrule
    \end{tabular}
\end{table}
Here, we measure the difference in performance between single task and multi-task learning.
For each of the 3 positive action types, we train a model to predict a single action type (10s Closeup, 10s Click, Repin), and then train a model to predict any of these 3 action types.
In Table \ref{tab:mtl-eval} we see the results: when we train on a specific action type, we maximize Recall@10 when treating only that action type as a positive label.
When training on all 3 action types together, we maximize the overall Recall@10, but perform slightly worse on each individual task than in a single-task setting.
For each task-specific eval, the multi-task performance is second to best, so we choose the multi-task training objective as a tradeoff between each objective, ensuring the final embedding does not strongly bias towards a specific task.

\subsubsection{Feature Ablations}
\begin{table}
    \centering
    \caption{Feature ablations, including all but one feature at a time. Removing any feature results in a drop in Recall@10.}
    \label{tab:feature-ablations}
    \begin{tabular}{lrr}
\toprule
Omitted Feature & P90 Coverage@10 & Recall@10 \\
\cmidrule(r){1-1}\cmidrule(l){2-3}
PinSage     &                0.0005 &       0.142 \\
Timestamp   &                0.050 &        0.210 \\
Surface     &                0.040 &        0.224 \\
Action Type &                0.042 &        0.226 \\
Duration    &                0.042 &        0.226 \\
Positional Encoding &        0.041 &        0.228 \\
None        &                0.042 &        0.229 \\
\bottomrule
\end{tabular}
\end{table}
In Table \ref{tab:feature-ablations}, we see the impact of each feature on the final model performance.
The two features that contribute noticeably to the final embedding are timestamp and the PinSage embedding.
Without PinSage, the model has no way of understanding content behind a user's action, and this is reflected both by low Recall@10, and very low global diversity, indicating that we retrieve a similar set of highly results for all users.
We see negative impact from removing each feature, so we choose to include all features in \longpsnew{}.

\subsubsection{Sequence Length}
\begin{figure}
    \centering
    \includegraphics[width=\columnwidth]{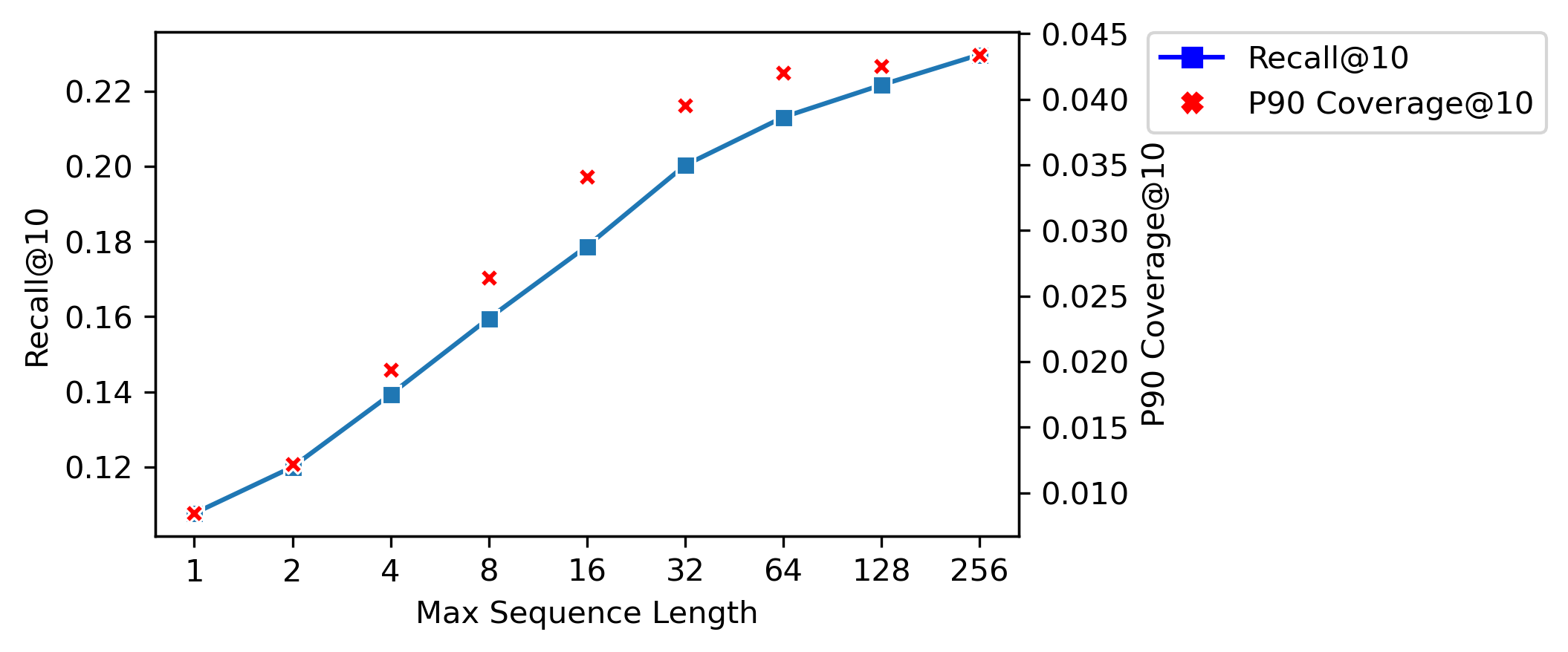}
    \caption{Max Sequence length vs Recall and Coverage. Longer sequences lead to better performance, but with diminishing returns.}
    \label{fig:seq-len}
\end{figure}
Figure \ref{fig:seq-len} shows the effect of sequence length on the model's performance.
We see approximately constant increases in both Recall@10 and global diversity when doubling sequence length up to around 32, but as sequence length increases we see diminishing returns.
In this work, we do not examine sequences longer than length 256, as such models require sacrifices either in terms of batch size or training resources.
A smaller batch size makes a comparison with shorter sequence models impossible, as the negative pool used for learning the embedding changes, and will require longer to train.
Using more machines (16 GPUs/2 machines for 512 sequence length, 32 GPUs/4 machines for 1024) allows longer sequence models to train, but decreases the number of possible parallel training runs.
With the ability to train fewer models in parallel, tuning modeling decisions become slower to make, so for \longpsnew{} we choose a sequence length of 256 in the final model.
\subsection{Ranking A/B Experiments}
We run several A/B experiments using it as a feature in ranking models to better understand how \longpsnew{} performs online.
\subsubsection{Homefeed}
\begin{table}
    \centering
    \caption{Online A/B experiment results replacing \psold{} with \psnew{} as a feature in our Homefeed ranking model. We see improvements in sitewide metrics.}
    \label{tab:pbty-exp}
    \hfill
    \begin{tabular}{lr}
        \toprule
        Metric & Lift  \\
        \midrule
        Time Spent & +1\% \\
        DAU & +0.4\% \\
        WAU & +0.12\% \\
        \bottomrule
    \end{tabular}
    \hfill
    \begin{tabular}{lr}
        \toprule
        Metric & Lift  \\
        \midrule
        Homefeed Repins & +7.5\% \\
        Homefeed Clickthroughs & +1\% \\
        Homefeed Close-ups & +6\% \\
        \bottomrule
    \end{tabular}
    \hfill
\end{table}
Our first comparison is in Pinterest's Homefeed ranking model, which helps determine the order in which content is shown to a user on Homefeed.
Previously this model used a weighted average of a user's top $k$ \longpsold{} embeddings as a feature.
In the experiment's enabled group, we replace this aggregation of \longpsold{} with the single \psnew{} embedding.
Both the control and enabled ranking models are trained on the same date range of data for fair comparisons.

Table \ref{tab:pbty-exp} shows key results from this experiment.
\psnew{} significantly improved engagement on Homefeed, and led to an increase in daily active users (DAUs) and weekly active users (WAUs).
We saw no regression in improvements over the course of several months after shipping the experiment.

\subsubsection{Ads}
To verify that this embedding is useful as a feature beyond use cases that it is explicitly trained on, we also run an A/B experiment adding \psnew{} to Ads ranking models (without replacing \longpsold{}).
Each primary surface (Homefeed, Related Pins, and Search) has a separate model dedicated to determining the order in which we show advertisements to users, so we experiment with each of them independently.
Overall, we see significant gains in engagement with ads on each sufrace, in terms of clickthrough rate (CTR) and long clickthrough rate (gCTR), which are shown in Table \ref{tab:ads-exp}.
\begin{table}
    \centering
    \caption{Online A/B experiment results adding \psnew{} as a feature to Ads ranking models. Each surface benefits signficantly from \longpsnew{}}
    \label{tab:ads-exp}
    \begin{tabular}{lrrr}
        \toprule
        Metric & Related Pins & Search & Homefeed  \\
        \cmidrule(r){1-1}\cmidrule(l){2-4}
        CTR & +7.1\% & +7.3\% & +10.0\%\\
        gCTR & +6.9\% & +5.2\% & +10.1\% \\
        \bottomrule
    \end{tabular}
\end{table}

\section{Conclusion}
In this work, we presented \longpsnew{}, a single, end-to-end learned embedding designed to be inferred in an offline setting, and to capture a user's interests over a multi-day time horizon.

In contrast to other work focused on modeling users based on their past actions, we do not focus directly on predicting a user's next engagement, but apply a novel loss function to capture a user's interests over a horizon of several days.
We show this training objective decreases the gap in performance between a model inferred in realtime and a model inferred once a day.
We also present detailed experiments to show the contribution of each component of our model to overall performance, demonstrating the effectiveness of multi-task learning and sampled softmax.

In the future, we plan to more thoroughly investigate performance of \longpsnew{} as a candidate generator, and include actions beyond Pin engagement as elements of the sequence of user actions, helping build an even more comprehensive representation of users.
\begin{acks}
The authors would like to thank
Jay Adams, Dhruvil Badani, Kofi Boakye, Haoyu Chen, Yi-ping Hsu, Haomiao Li, Yang Liu, Cosmin Negruseri, Yan Sun and Jiajing Xu
who contributed or supported us throughout this project.
\end{acks}
\bibliographystyle{ACM-Reference-Format}
\bibliography{sample-base}

\pagebreak
\appendix
\section{Information for Reproducibility}
\subsection{Timestamp Encoding}\label{appendix:timestamp}
We use 2 derived values in addition to the raw timestamp to represent time: the difference between the latest timestamp in the sequence and an action's timestamp, and the time gap between each two consecutive actions in the sequence, setting the last one to zero.
To encode timestamps, we modify Time2vec to use fixed periods, and apply a log transform of the raw time values
Specifically, given a timestamp $t$, and $P$ periods, $\{p_1, p_2, \ldots p_P \}$, we obtain $2P + 1$ features $r_1, \ldots, r_{2P + 1}$ by
\begin{align*}
r(t)_{2i - 1} &= \cos\left(\frac{2\pi t}{p_i} + \phi_{2i - 1}\right)\\
r(t)_{2i} &= \sin\left(\frac{2\pi t}{p_i} + \phi_{2i}\right),& i=1,\ldots,P \\
r(t)_{2P + 1} &= \log(t)
\end{align*}
where $\phi$ is a learned vector.
We select the periods manually, choosing to use $P_{\text{abs}}$ periods of real-life importance and fractions thereof: 0.25h, 0.5h, 0.75h, 1h, 2h, 4h, 8h, 16h, 1d, 7d, 28d, and 365d.

We encode relative time difference features using $P_{\text{rel}}=32$ evenly spaced periods on a log scale, ranging from one second to four weeks.
This assumes that it is more important for the model to be able to distinguish between short durations, such as ten seconds vs one minute, as compared to long durations, such as 10 days vs 11 days.

\subsection{Model Architecture}\label{appendix:model-arch}
Here we describe in more detail the transformer architecture we use.
We first construct the input matrix $A = \begin{pmatrix} A_{T} \cdots & A_{T - M + 1} \end{pmatrix}^\intercal \in \mathbb{R}^{M \times D_{\text{in}}}$ using the vector representations of the $M$ actions leading up to action $a_{T+1}$ as the sequence.
We first project this to the model's hidden dimension $H$ using a learnable matrix $W\in \mathbb{R}^{D_{\text{in}} \times H}$, then add a positional encoding $\mathrm{PE}\in \mathbb{R}^{M \times H}$.
This generates an input $V^{(0)} = A W + \mathrm{PE} \in \mathbb{R}^{M \times H}$ for the transformer.

Following this, we apply a standard transformer model, consisting of alternating 2-layer feedforward network (FFN) blocks and multi-head self attention (MHSA) blocks, where the hidden dimension of the feedforward network is four times the transformer hidden dimension.
In each MHSA block, we apply masking so a given output may only attend to current or previous elements of the sequence.

The model architecture can be described as follows:
\begin{equation}
  \label{eqn:transformer}
\begin{aligned}
    U^{(l)} &= V^{(l-1)} + \operatorname{MHSA}\left(\operatorname{LayerNorm}\left(V^{(l-1)}\right)\right) \\
    V^{(l)} &= U^{(l)} + \operatorname{FFN}\left(\operatorname{LayerNorm}\left(U^{(l)}\right)\right), &l = 1,\ldots,L
\end{aligned}
\end{equation}
After transforming inputs as described in Equation \ref{eqn:transformer}, we pass the final hidden state $V^{(L)} \in \mathbb{R}^{M \times H}$ through a two-layer MLP, then $L_2$ normalize the output.

The output MLP after the transformer is defined by \begin{equation*}
    W_2^\intercal \operatorname{GELU}(W_1^\intercal \operatorname{LayerNorm}(e) + b_2) + b_2
\end{equation*}
where $W_1 \in \mathbb{R}^{H \times 4H}, W_2 \in \mathbb{R}^{4H \times D}$, where $H$ is the transformer hidden dimension, $D$ is the embedding dimension, and $e\in \mathbb{R}^{256}$ is a single embedding.

This results in a set of embeddings $E = \begin{pmatrix} e_{1} & \cdots & e_{M} \end{pmatrix}^\intercal \in \mathbb{R}^{S \times D}$, where $D$ is the final embedding dimension.
We use $e_1$, the first row of $E$ and the most recent output, as the final user embedding.
In the case that a user does not have $M$ engagements, input sequences can be padded to length $M$, and positions that are padded can be masked in attention and loss computation, similarly to how they are treated in language modeling tasks.

\subsection{Mixed Negative Sampling Masking}\label{appendix:mns-masking}
\begin{figure}
    \centering%
    \includegraphics[width=\columnwidth]{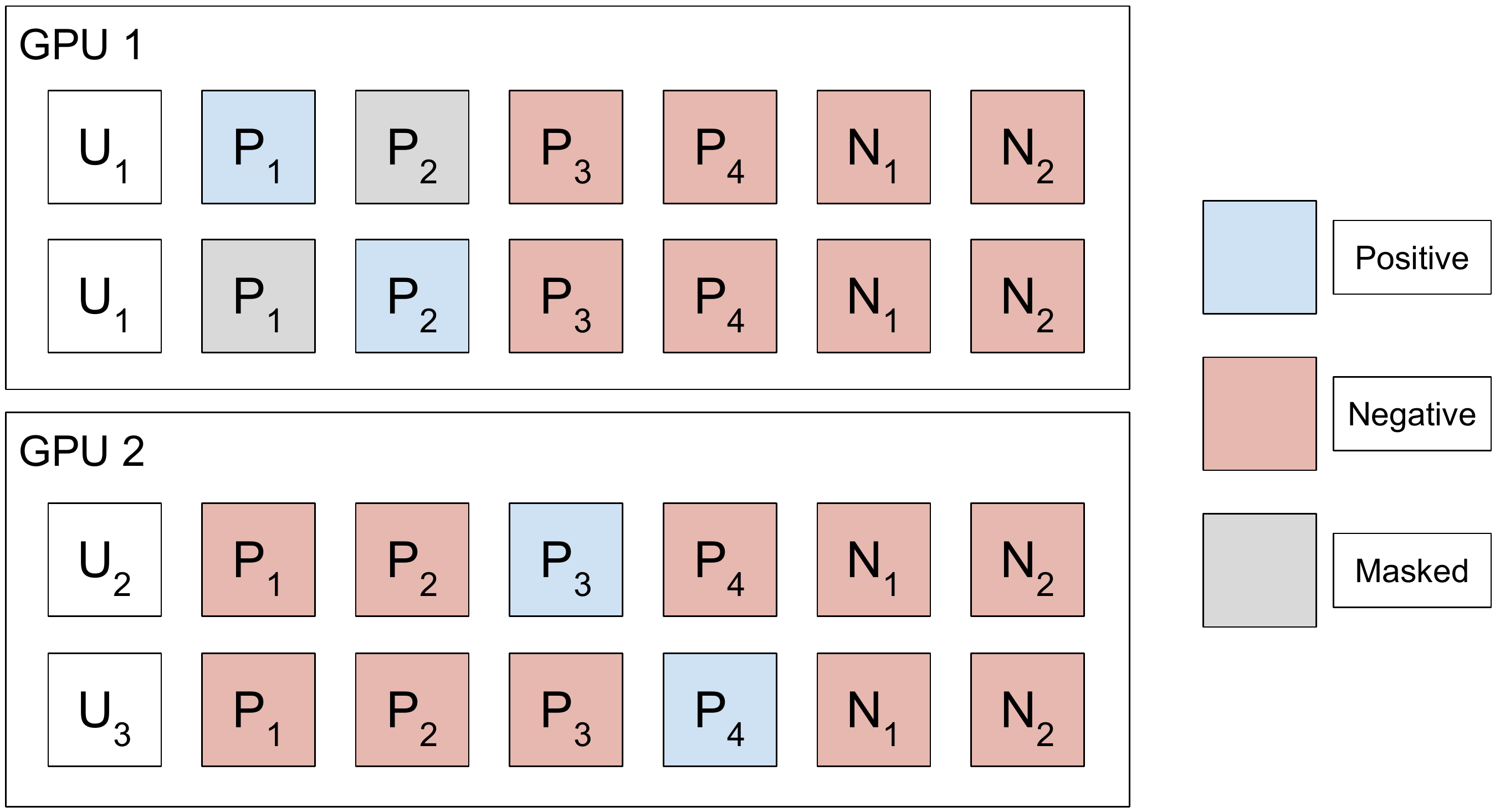} 
    \caption{Mixed negative sampling and masking. Further described in Section \ref{appendix:mns-masking} }
    \label{fig:loss_masking}
\end{figure}

In Figure \ref{fig:loss_masking} we depict mixed negative sampling with masking.
There are two user embeddings for user $U_1$ (potentially at different times) on GPU 1, and an embedding for $U_2$ and $U_3$ on GPU 2.
$U_1$ engaged with $P_1$ and $P_2$, $U_2$ engaged with $P_3$, and $U_3$ engaged with $P_4$.
$N_1$ and $N_2$ are random negatives.
When computing the loss, we treat each positive as a separate row, but mask $P_2$ in the first row and $P_1$ in the second, as they are both positive examples for $U_1$.
All four positives appear in both processes, as they are synchronized across GPUs before loss computation.
Each user per GPU gets equal weight in the final loss computation, so in this case user $U_1$ will be assigned twice the weight of $U_2$ or $U_3$.
In practice, a batch will contain many users, so weighting will be nearly uniform across all GPUs, even if it is not perfectly uniform.

In our experiments, we cap the number of in-batch negatives at 5000, and fix the number of random negatives to 8192.

\subsection{Architecture Ablations}
Here, we show the impact of varying model hyperparameters:
\subsubsection{Sequence Selection}
We have thoroughly explored removing weaker engagement from a user's history to generate a better embedding for users who engage often on Pinterest, but have seen no significantly positive results from sparsifying user sequences.

\subsubsection{Embedding Dimension}
\begin{figure}
    \centering
    \includegraphics[width=\columnwidth]{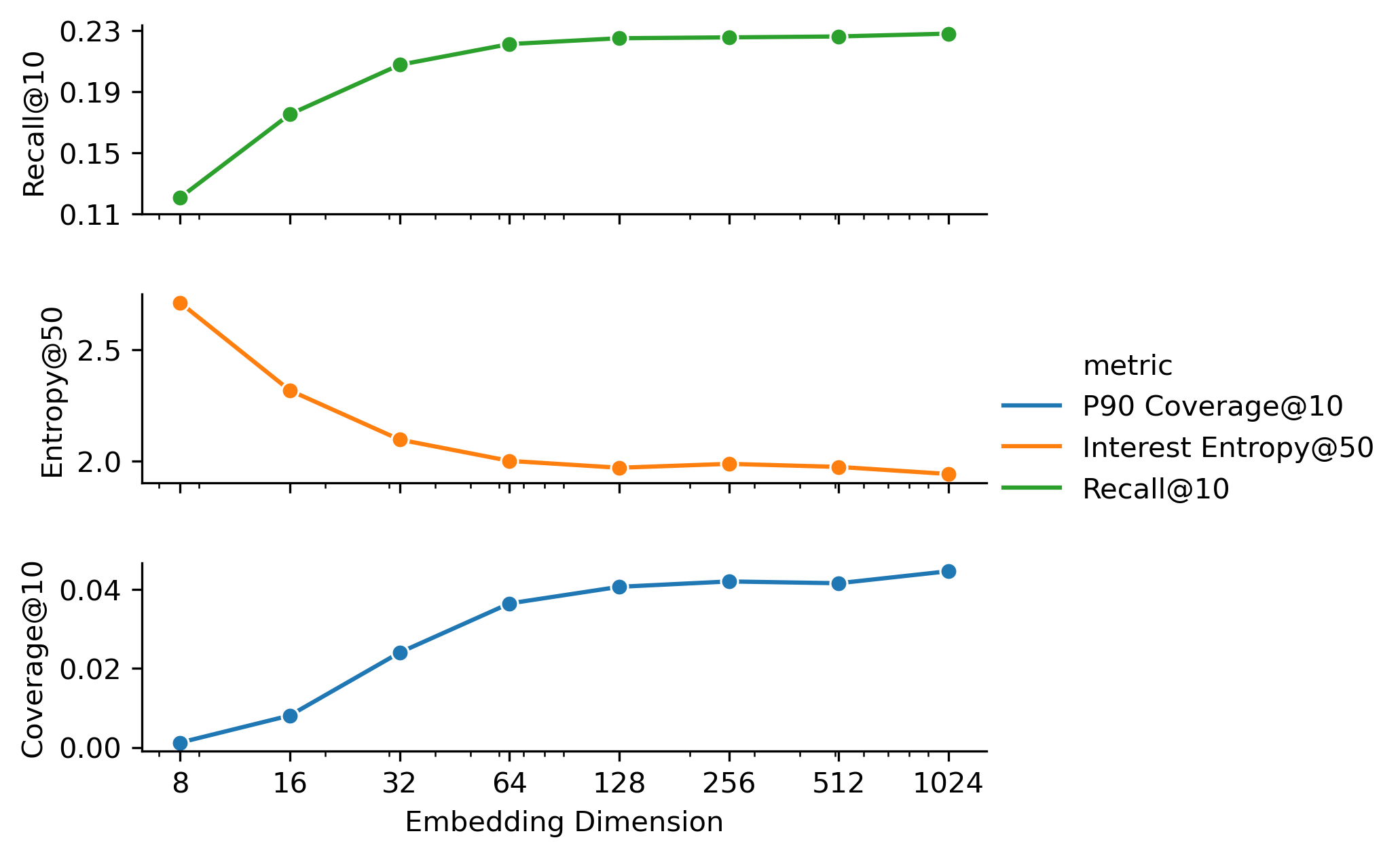}
    \caption{Recall and diversity vs embedding dimension. Smaller embeddings perform worse and are more likely to retrieve the same results for many users.}
    \label{fig:embedding-dim}
\end{figure}

In Figure \ref{fig:embedding-dim}, we show the effect of varying the size of our final embedding on overall performance.
We see diminishing improvements in Recall@10 as embedding dimension increases, especially beyond a 128d embedding.
We also see that at smaller dimensions, the embedding tends towards retrieving similar results for most users, likely implying a level of memorization of popularity.
As a consequence, there can be more diversity in each user's retrieved results at small dimensionalities, but because significant Recall@10 is sacrificed, this isn't a good tradeoff.
We choose to use a 256d embedding because it offers good offline metrics, and is the same size as most existing embedding features used in Pinterest's ranking models; the negligible increase in performance from increasing the embedding to 1024d is not worth quadrupling storage cost for most downstream use cases.

\subsubsection{Transformer Architecture}
\begin{table}
    \centering
    \caption{Recall vs Model Capacity. Larger models tend to perform better.}
    \label{tab:model-arch}
    \begin{tabular}{ccrr}
\toprule
\begin{tabular}{@{}c@{}}Num\\Layers\end{tabular} & \begin{tabular}{@{}c@{}}Hidden\\Dimension\end{tabular} & P90 Coverage@10 & Recall@10 \\
\cmidrule(r){1-2}\cmidrule(l){3-4}
2 & 256 &                0.0359 &             0.2189 \\
2 & 512 &                0.0388 &             0.2241 \\
2 & 768 &                0.0397 &             0.2246 \\
4 & 256 &                0.0366 &             0.2236 \\
4 & 512 &                0.0412 &             0.2240 \\
4 & 768 &                0.0426 &             0.2272 \\
6 & 256 &                0.0383 &             0.2233 \\
6 & 512 &                0.0400 &             0.2264 \\
6 & 768 &                0.0417 &             0.2293 \\
\bottomrule
    \end{tabular}
\end{table}

In Table \ref{tab:model-arch}, we show the effect of model capacity on final performance.
Larger models improve recall, both in terms of number of layers, and hidden size.
We do not see substantial changes when varying the number of heads used for multi-head self attention, so we hold this constant at 8 heads.

\subsubsection{Modification of SASRec}\label{appendix:sasrec}
\begin{table}
  \centering
  \caption{SASRec, Sampled Softmax vs Binary Cross Entropy loss. Sampled softmax performs significantly better than binary cross-entropy on our dataset.}
  \label{tab:sasrec-perf}
  \begin{tabular}{lrr}
    \toprule
    SASRec Loss & Recall@10 & P90 Coverage@10\\
    \cmidrule(r){1-1}\cmidrule(l){2-3}
Binary Cross-entropy            &  0.138   &   0.111 \\ %
\midrule
Sampled Softmax          &  0.181   &   0.056 \\ %
\midrule
\begin{tabular}{@{}l@{}}Sampled Softmax +\\equal $e_1$ loss weight\end{tabular}         &  0.198   &   0.048 \\
    \bottomrule
  \end{tabular}
\end{table}

In the original paper, SASRec \cite{kang2018self} model is trained based on a binary cross-entropy task, without any sample probability correction.
We make two modifications: (a) we give equal weight to the loss on $e_1$, the latest user embedding, and (b) we replace binary cross-entropy with sampled softmax.
In Table \ref{tab:sasrec-perf}, we show our modifications substantially improve recall.

\end{document}